\documentclass[12pt,twoside]{article}
\def\,{\mskip 3mu} \def\>{\mskip 4mu plus 2mu minus 4mu} \def\;{\mskip 5mu plus 5mu} \def\!{\mskip-3mu}
\def\dispmuskip{\thinmuskip= 3mu plus 0mu minus 2mu \medmuskip=  4mu plus 2mu minus 2mu \thickmuskip=5mu plus 5mu minus 2mu}
\def\textmuskip{\thinmuskip= 0mu                    \medmuskip=  1mu plus 1mu minus 1mu \thickmuskip=2mu plus 3mu minus 1mu}
\textmuskip
\def\beq{\dispmuskip\begin{equation}}    \def\eeq{\end{equation}\textmuskip}
\def\beqn{\dispmuskip\begin{displaymath}}\def\eeqn{\end{displaymath}\textmuskip}
\def\bqa{\dispmuskip\begin{eqnarray}}    \def\eqa{\end{eqnarray}\textmuskip}
\def\bqan{\dispmuskip\begin{eqnarray*}}  \def\eqan{\end{eqnarray*}\textmuskip}

\def\paradot#1{\vspace{1.3ex plus 0.5ex minus 0.5ex}\noindent{\bf{#1.}}}
\def\paranodot#1{\vspace{1.3ex plus 0.5ex minus 0.5ex}\noindent{\bf{#1}}}

\newtheorem{theorem}{Theorem}

\newtheorem{lemma}[theorem]{Lemma}
\newtheorem{proposition}[theorem]{Proposition}
\newtheorem{conjecture}[theorem]{Conjecture}

\newtheorem{definition}[theorem]{Definition}
\def\qed{\hspace*{\fill}\rule{1.4ex}{1.4ex}$\quad$\\}
\def\qedx{}

\newenvironment{proof}{\paradot{Proof}}{\qed}

\newenvironment{keywords}%
  {\centerline{\bf\small Keywords}\begin{quote}\small}%
  {\par\end{quote}\vskip 1ex}

\let\phi\varphi

\def\E{{\bf E}}

\def\odt{{\textstyle{1\over 2}}}
\def\odn{{\textstyle{1\over n}}}

\def\SetN{I\!\!N}
\def\C{{\cal C}}                        
\def\X{{\cal X}}                        
\def\O{{\cal O}}                        
\def\C{{\cal C}}                        

\def\eps{\varepsilon}                   

\begin{document}


\title{\vspace{-3ex}\normalsize\sc Technical Report \hfill IDSIA-13-06
\vskip 2mm\bf\Large\hrule height5pt \vskip 6mm
On Sequence Prediction for Arbitrary Measures
\vskip 6mm \hrule height2pt \vskip 5mm}
\author{{{\bf Daniil Ryabko} and {\bf Marcus Hutter}}\\[3mm]
\normalsize IDSIA, Galleria 2, CH-6928\ Manno-Lugano, Switzerland%
\thanks{This work was supported by the Swiss NSF grant 200020-107616.
}\\
\normalsize \{daniil,marcus\}@idsia.ch, \ http://www.idsia.ch/$^{_{_\sim}}\!$\{daniil,marcus\} }
\date{16 June 2006}
\maketitle
\vspace{-4ex}

\begin{abstract}
Suppose we are given two probability measures on the set of
one-way infinite finite-alphabet sequences and consider the
question when one of the  measures predicts the other, that is,
when conditional probabilities  converge (in a certain sense) when
one of the measures is chosen to generate the sequence. This
question may be considered a refinement of the problem of sequence
prediction in its most general formulation: for a given  class of
probability measures, does there exist a measure which predicts
all of the measures in the class? To address this problem, we find
some conditions on local absolute continuity which are sufficient
for prediction and which generalize several different notions
which are known to be sufficient for prediction. We also formulate
some open questions to outline a direction for finding the
conditions on classes of measures for which prediction is
possible.
\vspace{3ex}
\def\contentsname{\centering\normalsize Contents}
{\parskip=-2.5ex\tableofcontents}
\end{abstract}


\section{Introduction}

Let a sequence $x_t$, $t\in\SetN$ of letters from some finite
alphabet $\X$ be generated by some probability measure $\mu$.
Having observed the first $n$ letters $x_1,\dots,x_n$ we want to
predict what is the probability of the next letter being $x$, for
each $x\in\X$. This task is motivated by numerous applications ---
from weather forecasting and stock market prediction to data
compression.

If the measure $\mu$ is known completely then the best forecasts
one can make for the $(n+1)$st  outcome of a sequence
$x_1,\dots,x_n$ is $\mu$-conditional probabilities of  $x\in
\X$ given $x_1,\dots,x_n$. On the other hand, it is immediately
apparent that if nothing is known about the distribution $\mu$
generating the sequence then no prediction is possible, since for
any predictor there is a measure on which it errs (gives
inadequate probability forecasts) on every step. Thus one has to
restrict the attention to some class of measures. Laplace was
perhaps the first to address the question of sequence prediction,
his motivation being as follows: Suppose that we know that the Sun
has risen every day for 5000 years, what is the probability that
it will rise tomorrow? He suggested to assume that the probability
that the Sun rises is the same every day and the trials are
independent of each other. Thus Laplace considered the task of
sequence prediction when the true generating measure belongs to
the family of Bernoulli i.i.d.\ measures  with binary alphabet
$\X=\{0,1\}$. The predicting measure suggested by Laplace was
$\rho_L(x_{n+1}=1|x_1,\dots,x_n)=\frac{k+1}{n+2}$ where $k$ is the
number of 1s in $x_1,\dots,x_n$. The conditional probabilities of
Laplace's measure $\rho_L$ converge to the true conditional
probabilities $\mu$-almost surely under any Bernoulli i.i.d
measure $\mu$. This approach generalizes to the problem of
predicting any finite-memory (e.g.\ Markovian) measure. Moreover,
in \cite{BRyabko:88} a measure $\rho_R$ was constructed for
predicting an arbitrary stationary measure. The conditional
probabilities of $\rho_R$ converge to the true ones {\em on
average}, where average is taken over time steps (that is, in
Cesaro sense), $\mu$-almost surely for any stationary measure
$\mu$. However, as it was shown in the same work, there is no
measure for which conditional probabilities converge to the true
ones $\mu$-a.s.\ for every stationary $\mu$. Thus we can see that
already for the problem of predicting outcomes of a stationary
measure two criteria of prediction arise: prediction in the
average (or in Cesaro sense) and prediction on each step, and the
solution exists only for the former problem.

But what if the measure generating the sequence is not stationary?
A different assumption one can make is that the measure $\mu$
generating the sequence is computable. Solomonoff
\cite[Eq.(13)]{Solomonoff:64} suggested a measure $\xi$ for
predicting any computable probability measure. The key observation
here is that the class of all computable probability measures is
countable; let us denote it by $(\nu_i)_{i\in\SetN}$. A Bayesian
predictor $\xi$ for a countable class of measures
$(\nu_i)_{i\in\SetN}$  is constructed as follows:
$\xi(A)=\sum_{i=1}^\infty w_i\nu_i(A)$ for any measurable set A,
where the weights $w_i$ are positive  and sum to one\footnote{It
is not necessary for prediction that the weights sum to one. In
\cite{Solomonoff:78} and \cite{Zvonkin:70} $w_i=2^{-K(i)}$
where $K$ stands for the prefix Kolmogorov complexity, and so the
weights do not sum to 1. Further, the $\nu$ and $\xi$ are only
semi-measures.}. The best predictor for a measure $\mu$ is the
measure $\mu$ itself. The Bayesian predictor simply takes the
weighted average of the predictors for all measures in the class
--- for countable classes this is possible. It was shown by
Solomonoff \cite{Solomonoff:78} that $\xi$-conditional
probabilities converge to $\mu$-conditional probabilities almost
surely for any computable measure $\mu$. In fact this is a special
case of a more general (though without convergence rate) result of
Blackwell and Dubins \cite{Blackwell:62} which states that if a
measure $\mu$ is absolutely continuous with respect to a measure
$\rho$ then $\rho$ converges to $\mu$ in total variation
$\mu$-almost surely. Convergence in total variation means
prediction in a very strong sense~--- convergence of conditional
probabilities of arbitrary events (not just the next outcome), or
prediction with arbitrary fast growing horizon. Since for $\xi$ we
have $\xi(A)\ge w_i\nu_i(A)$ for every measurable set $A$ and for
every $\nu_i$, each $\nu_i$ is absolutely continuous with respect
to $\xi$.

Thus the problem of sequence prediction for certain  classes of
measures (such as the class of all stationary measures or the
class of all computable measures) was often addressed in the
literature. Although the mentioned classes of measures are
sufficiently interesting, it is often hard to decide in
applications with which assumptions does a problem at hand comply;
not to mention such practical issues as that a predicting measure
for all computable measures is necessarily non-computable itself.
Moreover, to be able to generalize the solutions of the sequence
prediction problem to such problems as active learning, where
outcomes of a sequence may depend on actions of the predictor, one
has to understand better under which conditions  the problem of
sequence prediction is solvable. In particular, in active learning,
the stationarity assumption does not seem to be applicable (since
the predictions are non-stationary), although, say, the Markov
assumption is often applicable and is extensively studied. Thus, we
formulate the following general questions which we start to
address in the present work:

\paradot{General motivating questions}
For which classes of measures is sequence prediction possible?
Under which conditions does a measure $\rho$ predict a measure
$\mu$?

As we have seen, these questions have many facets, and in
particular there are many criteria of prediction to be considered,
such as almost sure convergence of conditional probabilities,
convergence in average, etc. Extensive as the literature on
sequence prediction is, these questions in their full generality
have not received much attention. One line of research which
exhibits this kind of generality consists in extending the result
of Blackwell and Dubins mentioned above, which states that if
$\mu$ is absolutely continuous with respect to $\rho$, then $\rho$
predicts $\mu$ in total variation distance. In \cite{Jackson:99} a
question of whether, given a class of measures $\C$ and a
prior (``meta''-measure) $\lambda$ over this class of measures, the
conditional probabilities of a Bayesian mixture of the class
$\C$ w.r.t. $\lambda$ converge to the true
$\mu$-probabilities (weakly merge, in terminology of
\cite{Jackson:99}) for $\lambda$--almost any measure $\mu$ in
$\C$. This question can be considered solved, since the
authors provide necessary and sufficient conditions on the
measure given by the mixture of the class $\C$ w.r.t.
$\lambda$ under which prediction is possible. The major difference
from the general  questions we posed above is that we do not wish
to assume that we have a measure on our class of measures. For
large (non-parametric) classes of measures it may not be intuitive
which measure over it is natural; rather, the question is  whether
a ``natural'' measure which can be used for prediction exists.

To address the general questions posed, we start with the
following observation. As it was mentioned, for a Bayesian mixture
$\xi$ of a countable class of measures $\nu_i$, $i\in\SetN$, we
have $\xi(A)\ge w_i\nu_i(A)$ for any $i$ and any measurable set
$A$, where $w_i$ is a constant. This condition is stronger than
the assumption of absolute continuity and is sufficient for
prediction in a very strong sense. Since we are willing to be
satisfied with prediction in a weaker sense (e.g.\ convergence of
conditional probabilities), let us make a weaker assumption: Say
that {\em a measure $\rho$ dominates a measure $\mu$ with
coefficients $c_n>0$} if
\beq\label{eq:dom}
  \rho(x_1,\dots,x_n) \;\geq\; c_n \mu(x_1,\dots,x_n)
\eeq
for all $x_1,\dots,x_n$.

\paranodot{The first concrete question}
we pose is, under what conditions on $c_n$ does (\ref{eq:dom})
imply that $\rho$ predicts $\mu$? Observe that if
$\rho(x_1,\dots,x_n)>0$ for any $x_1,\dots,x_n$ then any measure
$\mu$ is {\em locally} absolutely continuous with respect to
$\rho$ (that is, the measure $\mu$ restricted to the first $n$
trials $\mu|_{\X^n}$ is absolutely continuous w.r.t.
$\rho|_{\X^n}$ for each $n$), and moreover, for any measure $\mu$
some constants $c_n$ can be found that satisfy (\ref{eq:dom}). For
example, if $\rho$ is Bernoulli i.i.d.\ measure with parameter
$\odt$ and $\mu$ is any other measure, then (\ref{eq:dom}) is
(trivially) satisfied with $c_n=2^{-n}$. Thus we know that if
$c_n\equiv c$ then $\rho$ predicts $\mu$ in a very strong sense,
whereas exponentially decreasing $c_n$ are not enough for
prediction. Perhaps somewhat surprisingly, we will show that
dominance with any subexponentially decreasing coefficients is
sufficient for prediction, in a weak sense of convergence of
expected averages. Dominance with any polynomially decreasing
coefficients, and also with coefficients decreasing (for example)
as $c_n=\exp(-\sqrt{n}/\log n)$, is sufficient for (almost sure)
prediction on average (i.e.\ in Cesaro sense). However, for
prediction on every step we have a negative result: for any
dominance coefficients that go to zero there exists a pair of
measures $\rho$ and $\mu$ which satisfy~(\ref{eq:dom}) but $\rho$
does not predict $\mu$ in the sense of almost sure convergence of
probabilities. Thus the situation is similar to that for
predicting any stationary measure: prediction is possible in the
average but not on every step.

Note also that for Laplace's measure $\rho_L$ it can be shown that
$\rho_L$ dominates any i.i.d.\ measure $\mu$ with linearly
decreasing coefficients $c_n={1\over n+1}$; a generalization of
$\rho_L$ for predicting all measures with memory $k$ (for a given
$k$) dominates them with polynomially  decreasing coefficients.
Thus dominance with decreasing coefficients generalizes (in a
sense) predicting countable classes of measures (where we have
dominance with a constant), absolute continuity (via local
absolute continuity), and predicting i.i.d.\ and finite-memory
measures.

Another way to look for generalizations  is as follows. The Bayes
mixture $\xi$, being a sum of countably many measures
(predictors), possesses some of their predicting properties. In
general, which predictive properties are preserved under
summation?  In particular, if we have two predictors $\rho_1$ and
$\rho_2$ for two classes of measures, we are interested in the
question whether $\odt(\rho_1+\rho_2)$ is a predictor for the
union of the two classes. An answer to this question would improve
our understanding of how far a class of measures for which a
predicting measure exists can be extended without losing this
property.

\paranodot{{\rm Thus,} the second question}
we consider is the following: suppose that a measure $\rho$
predicts $\mu$ (in some weak sense), and let $\chi$ be some other
measure (e.g.\ a predictor for a different class of measures). Does
the measure $\rho'=\odt(\rho+\chi)$ still predict $\mu$?
That is, we ask to which prediction quality criteria does the idea
of taking a Bayesian sum generalize.  Absolute continuity is
preserved under summation along with it's (strong) prediction
ability. It was mentioned in \cite{BRyabko:06} that prediction in
the (weak) sense of convergence of expected averages of
conditional probabilities is preserved under summation. Here we
find that several stronger notions of prediction are not preserved
under summation.

Thus we address the following two questions. Is dominance with
decreasing coefficients sufficient for prediction in some sense,
under some  conditions on the coefficients? And, if a measure
$\rho$ predicts a measure $\mu$ in some sense, does the measure
$\odt(\rho+\chi)$ also predict $\mu$ in the same sense, where
$\chi$ is an arbitrary measure? Considering different criteria of
prediction (a.s.\ convergence of conditional probabilities, a.s.\
convergence of averages, etc.) in the above two questions we
obtain not two but many different questions, some of which we
answer in the positive and some in the negative,   yet some are
left open.

\paradot{Contents}
The paper is organized as follows. Section~\ref{sec:def}
introduces necessary notation and measures of divergence of
probability measures. Section~\ref{sec:dom} addresses the question
of whether dominance with decreasing coefficients is sufficient
for prediction, while in Section~\ref{sec:sum} we consider the
question of summing a predictor with an arbitrary measure. Both
sections~\ref{sec:dom} and~\ref{sec:sum} also propose some open
questions and directions for future research. In
Section~\ref{sec:misc} we discuss some interesting special cases
of the questions considered, and also some related problems.

\section{Notation and Definitions}\label{sec:def}

We consider processes on the set of  one-way infinite sequences
$\X^\infty$ where $\X$ is a finite set (alphabet).
In the examples we will often assume $\X=\{0,1\}$. The notation
$x_{1:n}$ is used for $x_1,\dots,x_n$ and $x_{<n}$ for
$x_1,\dots,x_{n-1}$, $x_t\in\X$. The symbol $\mu$ is reserved for
the ``true'' measure generating examples. We use  $\E_\nu$ for
expectation with respect to a measure $\nu$ and simply $\E$ for
$\E_\mu$ (expectation with respect to the ``true'' measure
generating examples).

For two measures $\mu$ and $\rho$ define the following measures of divergence.
\begin{itemize}\itemindent=1ex
\item[($d$)]{ Kullblack-Leibler (KL) divergence}\\ $\displaystyle  d_n(\mu,\rho|x_{<n})=\sum_{x\in\X}\mu(x_n=x|x_{<n})\log\frac{\mu(x_n=x|x_{<n})}{\rho(x_n=x|x_{<n})}$,
\item[($\bar d$)]{ average KL divergence} $\displaystyle  \bar d_n(\mu,\rho|x_{1:n})={1\over n} \sum_{t=1}^n d_t(\mu,\rho|x_{<n})$,
\item[($a$)]{ absolute distance} $\displaystyle a_n(\mu,\rho|x_{<n})=\sum_{x\in\X}|\mu(x_n=x|x_{<n})-\rho(x_n=x|x_{<n})|$,
\item[($\bar a$)]{ average absolute distance} $\displaystyle \bar a_n(\mu,\rho|x_{1:n})={1\over n}\sum_{t=1}^n a_t(\mu,\rho|x_{<n})$.
\end{itemize}

\begin{definition}[Convergence concepts]\label{def:conv}
We say that $\rho$ predicts $\mu$
\begin{itemize}\itemindent=2ex
\item[$(d)$\ ]      in {KL divergence} if $d_n(\mu,\rho|x_{<n})\rightarrow0$  $\mu$-a.s., \vspace{1mm}
\item[$(\bar d)$\ ] in {average KL divergence} if $\bar d_n(\mu,\rho|x_{1:n})\rightarrow 0$ $\mu$-a.s., \vspace{1mm}
\item[$(\E\bar d)$] in {expected average KL divergence} if $\E_\mu\bar d_n(\mu,\rho|x_{1:n})\rightarrow 0$, \vspace{1mm}
\item[$(a)$\ ]      in {absolute distance} if $a_n(\mu,\rho|x_{<n})\rightarrow0$  $\mu$-a.s., \vspace{1mm}
\item[$(\bar a)$\ ] in {average absolute distance} if $\bar a_n(\mu,\rho|x_{1:n})\rightarrow 0$ $\mu$-a.s., \vspace{1mm}
\item[$(\E\bar a)$] in {expected average absolute distance} if $\E_\mu\bar a_n(\mu,\rho|x_{1:n})\rightarrow 0$.
\end{itemize}
\end{definition}

The argument $x_{1:n}$ will be often left implicit in our
notation. A measure $\rho$ converges to $\mu$ in {\em total
variation} $(tv)$ if $\sup_{A\subset \sigma(\bigcup_{t=n}^\infty
\X^t)}|\mu(A|x_{<n}) - \rho(A|x_{<n})|\rightarrow0$ $\mu$-almost
surely. Some other measures of prediction ability are considered
in Section~{\ref{sec:misc}}. The following implications hold
(and are complete and strict):
\beqn
\begin{array}{ccccccc}
     &             & d & \Rightarrow & \bar d &             & \E\bar d \\
     &         & \Downarrow  &   & \Downarrow &             & \Downarrow \\
  tv & \Rightarrow & a & \Rightarrow & \bar a & \Rightarrow & \E\bar a \\
\end{array}%
\eeqn
to be understood as e.g.: if $\bar d_n\to 0$ a.s.\ then $\bar a_n\to 0$
a.s, or, if $\E\bar d_n\to 0$ then $\E\bar a_n\to 0$. The
horizontal implications $\Rightarrow$ follow immediately from the
definitions, and the $\Downarrow$ follow from the following Lemma:

\begin{lemma}[\boldmath $a^2\leq 2d$]\label{th:da}
For all measures $\rho$ and $\mu$ and sequences $x_{1:\infty}$ we
have: $a_t^2\leq 2d_t$ and $\bar a_n^2\leq 2\bar d_n$ and $(\E\bar
a_n)^2\leq 2\E\bar d_n$.
\end{lemma}

\begin{proof}
Pinsker's inequality \cite[Lem.3.11$a$]{Hutter:04uaibook} implies
$a_t^2\leq 2d_t$. Using this and Jensen's inequality for the average
$\odn\sum_{t=1}^n [...]$ we get
\beqn
  2\bar d_n
  \;=\; {1\over n}\sum_{t=1}^n 2d_t
  \;\geq\;   {1\over n}\sum_{t=1}^n a_t^2
  \;\geq\;  \left({1\over n}\sum_{t=1}^n a_t\right)^2
  \;=\; \bar a_n^2
\eeqn
Using this and Jensen's inequality for the expectation $\E$ we get
$2\E\bar d_n\geq \E\bar a_n^2 \geq (\E\bar a_n)^2$.
\qedx
\end{proof}

\section{Dominance with Decreasing Coefficients}\label{sec:dom}

First we consider the question whether property (\ref{eq:dom}) is
sufficient for prediction.

\begin{definition}[Dominance]\label{def:dom}
We say that a measure $\rho$ dominates a measure $\mu$
with coefficients $c_n>0$ iff
\beqn
 \rho(x_{1:n}) \;\geq\; c_n \mu(x_{1:n}).
\eeqn
\end{definition}

Suppose that $\rho$ dominates $\mu$ with decreasing coefficients
$c_n$. Does $\rho$ predict $\mu$ in (expected, expected average)
KL divergence (absolute distance)?
First let us give an example.

\begin{proposition}[Dominance of Laplace's measure]\label{prop:Laplace}
Let $\rho_L$ be the Laplace measure, given by
$\rho_L(x_{n+1}=a|x_{1:n})=\frac{k+1}{n+|\X|}$ for any $a\in\X$ and any
$x_{1:n}\in\X^n$, where $k$ is the number of occurrences of $a$
in $x_{1:n}$. Then
\beqn
  \rho_L(x_{1:n}) \;\geq\; \frac{n!}{(n+|\X|-1)!} \; \mu({x_{1:n}})
\eeqn
for any measure $\mu$ which generates independently and
identically distributed symbols. This bound is sharp.
\end{proposition}

\begin{proof}
We will only give the proof for $\X=\{0,1\}$, the general case is
analogous. To calculate $\rho_L(x_{1:n})$ observe that it only
depends on the number of 0s and 1s in $x_{1:n}$ and not on their
order. Thus we compute $\rho_L(x_{1:n})=\frac{k!(n-k)!}{(n+1)!}$
where $k$ is the number of 1s. For any measure $\mu$ such that
$\mu(x_n=1)=p$ for some $p\in[0,1]$ independently for all $n$, and
for Laplace measure $\rho_L$ we have
\bqan
  \frac{\mu(x_{1:n})}{\rho_L(x_{1:n})}
  & = & \frac{(n+1)!}{k!(n-k)!}p^k(1-p)^{n-k} \\
  & = & (n+1){n\choose k}p^k(1-p)^{n-k} \\
  & \le & (n+1) \sum_{k=1}^n{n\choose k}p^k(1-p)^{n-k}=n+1,
\eqan
for any $n$-letter word $x_1,\dots,x_n$  where $k$ is the number of 1s in it.
The bound is attained when $p=1$, so that $k=n$, $\mu(x_{1:n})=1$,
and $\rho_L(x_{1:n})=\frac{1}{n+1}$. \qedx
\end{proof}

Thus for Laplace's measure $\rho_L$ and binary $\X$ we have
$c_n=\O(\frac{1}{n})$. As it was mentioned in the
introduction, in general, exponentially decreasing  coefficients
$c_n$ are not sufficient for prediction, since (\ref{eq:dom}) is
satisfied with $\rho$ being a Bernoulli i.i.d.\ measure and $\mu$
any other measure. On the other hand, the following proposition
shows that in a weak sense of convergence in expected average KL
divergence (or absolute distance) the property~(\ref{eq:dom}) with
subexponentially decreasing $c_n$ is sufficient. We also remind
that  if $c_n$ are bounded from below then prediction in the
strong sense of total variation is possible.

\begin{theorem}[\boldmath $\E\bar d\to 0$ and $\E\bar a\to 0$]\label{th:dom}
Let $\mu$ and $\rho$ be two measures on $\X^\infty$ and suppose that
$
 \rho(x_{1:n})\ge c_n \mu(x_{1:n})
$
for any $x_{1:n}$, where $c_n$ are positive constants satisfying
$\frac{1}{n}\log c_n^{-1}\rightarrow0$. Then $\rho$ predicts
$\mu$ in expected average KL divergence $\E_\mu \bar
d_n(\mu,\rho)\rightarrow0$ and in expected average absolute
distance $\E_\mu \bar a_n(\mu,\rho)\rightarrow0$.
\end{theorem}

The proof of this  proposition is based on the same idea as the
proof of convergence of Solomonoff predictor to any of its
summands in \cite{BRyabko:88}, see also \cite{Hutter:04uaibook}.

\begin{proof}
For convergence in average expected KL divergence we have
\bqan
  \E_\mu\bar d_n(\mu,\rho)
  & = & \frac{1}{n}\E\sum_{t=1}^n \sum_{x_t\in\X} \mu(x_t|x_{<t})\log \frac {\mu(x_t|x_{<t})}{\rho(x_t|x_{<t})}
  \;=\; \frac{1}{n}\sum_{t=1}^n\E\E^t\log \frac {\mu(x_t|x_{<t})}{\rho(x_t|x_{<t})}
\\
  & = & \frac{1}{n}\E\log\prod_{t=1}^n\frac {\mu(x_t|x_{<t})}{\rho(x_t|x_{<t})}
  \;=\; \frac{1}{n}\E \log \frac {\mu(x_{1:n})}{\rho(x_{1:n})}
  \;\le\; \frac{1}{n}  \log c_n^{-1}\rightarrow0,
\eqan
where $\E^t$ stands for the $\mu$-expectation over $x_t$
conditional on $x_{<t}$.

The statement for expected average distance follows from this and
Lemma \ref{th:da}. \qedx
\end{proof}

With a stronger condition on $c_n$ prediction in average KL
divergence can be established.

\begin{theorem}[\boldmath $\bar d\to 0$ and $\bar a\to 0$]\label{th:dom2}
Let $\mu$ and $\rho$ be two measures on $\X^\infty$ and suppose that
$
 \rho(x_{1:n})\ge c_n \mu(x_{1:n})
$
for every $x_{1:n}$,
where $c_n$ are positive constants satisfying
\beq\label{eq:domsum}
 \sum_{n=1}^{\infty} \frac {(\log c_n^{-1})^2}{n^2} \;<\; \infty.
\eeq
Then $\rho$ predicts $\mu$  in  average KL divergence $\bar
d_n(\mu,\rho)\rightarrow0$ $\mu$-a.s.\ and in  average absolute
distance $\bar a_n(\mu,\rho)\rightarrow0$ $\mu$-a.s.
\end{theorem}

In particular, the condition~(\ref{eq:domsum}) on the coefficients
is satisfied for polynomially decreasing coefficients, or for
$c_n=\exp(-\sqrt{n}/\log n)$.

\begin{proof}
Again the second statement (about absolute distance) follows from
the first one and Lemma~\ref{th:da}, so that we only have to prove
the statement about KL divergence.

Introduce the symbol $\E^n$ for $\mu$-expectation over $x_n$
conditional on $x_{<n}$. Consider random variables
$l_n=\log\frac{\mu(x_n|x_{<n})}{\rho(x_n|x_{<n})}$ and $\bar l_n=
{1\over n} \sum_{t=1}^n l_t$. Observe that $d_n=\E^n l_n$,
so that the random variables $m_n=l_n-d_n$ form a martingale
difference sequence (that is, $\E^n m_n=0$). Let also $\bar m_n=
{1\over n} \sum_{t=1}^n m_t$. We will show that $\bar
m_n\rightarrow0$ $\mu$-a.s.\ and $\bar l_n\rightarrow0$ $\mu$-a.s.
which implies $\bar d_n\rightarrow 0$
$\mu$-a.s.

Note that
\beqn
 \bar l_n \;=\; {1\over n}\log \frac{\mu(x_{1:n})}{\rho(x_{1:n})}
 \;\leq\; \frac{\log c_n^{-1}}{n} \;\rightarrow\; 0.
\eeqn
Thus to show that $\bar l_n$ goes to $0$ we need to bound it from
below. It is easy to see that $n\bar l_n$ is ($\mu$-a.s.) bounded
from below by a constant, since $\frac{\rho (x_{1:n})}{\mu
(x_{1:n})}$ is a $\mu$-martingale whose expectation is 1, and so
it converges to a finite limit $\mu$-a.s.\ by Doob's submartingale
convergence theorem, see e.g.\ \cite[p.508]{Shiryaev:96}.

Next we will show that $\bar m_n\rightarrow0$ $\mu$-a.s.
 We have
\bqan
  m_n & = & \log\frac{\mu(x_{1:n})}{\rho(x_{1:n})}
          - \log\frac{\mu(x_{<n})}{\rho(x_{<n})}
      - \E^n\log\frac{\mu(x_{1:n})}{\rho(x_{1:n})}
      + \E^n\log\frac{\mu(x_{<n})}{\rho(x_{<n})}
 \\
      & = & \log\frac{\mu(x_{1:n})}{\rho(x_{1:n})}
      - \E^n\log\frac{\mu(x_{1:n})}{\rho(x_{1:n})}.
\eqan
Let $f(n)$ be some function monotonically increasing to infinity such that
\beq\label{eq:f}
  \sum_{n=1}^\infty\frac{(\log c_n^{-1}+f(n))^2}{n^2} \;<\; \infty
\eeq
(e.g.\ choose $f(n)=\log n$ and exploit
$(\log c_n^{-1}+f(n))^2 \leq 2(\log c_n^{-1})^2+2f(n)^2$ and (\ref{eq:domsum}).)
For a sequence of random variables  $\lambda_n$ define
\beqn
  (\lambda_n)^{+(f)} \;=\; \left\{ \begin{array}{ll} \lambda_n & \mbox{ if }
    \lambda_n\ge - f(n) \\ 0 & \mbox{ otherwise }\end{array}\right.
\eeqn
and $\lambda_n^{-(f)}=\lambda_n - \lambda_n^{+(f)}$.
Introduce also
\beqn
  m^+_n \;=\; \left (\log \frac{\mu(x_{1:n})}{\rho(x_{1:n})}\right)^{+(f)}
         - \E^n\left(\log\frac{\mu(x_{1:n})}{\rho(x_{1:n})}\right)^{+(f)},
\eeqn
$m_n^-=m_n-m_n^+$ and the averages $\bar m^+_n$ and $\bar m_n^-$.
Observe that $m_n^+$ is a martingale difference sequence. Hence to
establish the convergence $\bar m^+_n\rightarrow0$ we can use the
martingale strong law of large numbers \cite[p.501]{Shiryaev:96},
which states that, for a martingale difference sequence
$\lambda_n$, if $\E(n\bar \lambda_n)^2<\infty$ and
$\sum_{n=1}^\infty \E \lambda_n^2/n^2<\infty$ then $\bar
\lambda_n\rightarrow0$ a.s. Indeed, for $m^+_n$ the first
condition is trivially satisfied (since the expectation in
question is a finite sum of finite numbers), and the second
follows from the fact that $|m_n^+|\le \log c_n^{-1}+f(n)$
and~(\ref{eq:f}).

Furthermore, we have
\beqn
  m_n^- \;=\; \left(\log \frac{\mu(x_{1:n})}{\rho(x_{1:n})}\right)^{-(f)}
        - \E^n\left(\log\frac{\mu(x_{1:n})}{\rho(x_{1:n})}\right)^{-(f)}.
\eeqn
As it was mentioned before, $\log
\frac{\mu(x_{1:n})}{\rho(x_{1:n})}$ converges $\mu$-a.s.\ either
to (positive) infinity or to a finite number. Hence $\left(\log
\frac{\mu(x_{1:n})}{\rho(x_{1:n})}\right)^{-(f)}$ is non-zero
only a  finite  number of times, and so  its average goes to zero.
To see that $\E^n\left(\log
\frac{\mu(x_{1:n})}{\rho(x_{1:n})}\right)^{-(f)}\rightarrow0$ we
write
\bqan
  \E^n\left(\log \frac{\mu(x_{1:n})}{\rho(x_{1:n})}\right)^{-(f)}
  & =& \sum_{x_n\in\X}\mu(x_n|x_{<n}) \left( \log \frac{\mu(x_{<n})}{\rho(x_{<n})}
        + \log\frac{\mu(x_n|x_{<n})}{\rho(x_n|x_{<n})}\right)^{-(f)}
\\
  & \ge & \sum_{x_n\in\X}\mu(x_n|x_{<n}) \left( \log \frac{\mu(x_{<n})}{\rho(x_{<n})}
        + \log\mu(x_n|x_{<n})\right)^{-(f)}
\eqan
and note that the first term in brackets is bounded from below,
and so for the sum in brackets to be less than $-f(n)$ (which is
unbounded) the second term $\log\mu(x_n|x_{<n})$ has to go to
$-\infty$, but then the expectation goes to zero since
$\lim_{u\rightarrow0} u\log u=0$.

Thus we conclude that $\bar m_n^-\rightarrow0$ $\mu$-a.s., which
together with $\bar m_n^+\rightarrow0$ $\mu$-a.s.\ implies $\bar
m_n\rightarrow0$ $\mu$-a.s., which, finally, together with $\bar
l_n\rightarrow0$ $\mu$-a.s.\ implies $\bar d_n\rightarrow0$
$\mu$-a.s. \qedx
\end{proof}

However, no form of dominance with decreasing coefficients is
sufficient for prediction in absolute distance or KL divergence,
as the following negative result states.

\begin{proposition}[\boldmath $d\not\to 0$ and $a\not\to 0$]\label{th:nodom}
For each sequence of positive numbers $c_n$ that goes to 0 there
exist measures $\mu$ and $\rho$ and a number $\epsilon>0$ such
that
$
 \rho(x_{1:n})\ge c_n \mu(x_{1:n})
$
for all $x_{1:n}$, yet $a_n(\mu,\rho|x_{1:n})>\epsilon$ and
$d_n(\mu,\rho|x_{1:n})>\epsilon$ infinitely often $\mu$-a.s.
\end{proposition}

\begin{proof}
Let $\mu$ be concentrated on the sequence $11111\dots$ (that is
$\mu(x_n=1)=1$ for all $n$), and let $\rho(x_n=1)=1$ for all $n$
except for a subsequence of steps $n=n_k$, $k\in\SetN$ on which
$\rho(x_{n_k}=1)=1/2$ independently of each other. It is easy to
see that choosing $n_k$ sparse enough we can make
$\rho(1_1\dots1_n)$ decrease arbitrary slowly; yet
$|\mu(x_{n_k})-\rho(x_{n_k})|=1/2$ for all $k$. \qedx
\end{proof}

Thus for the first question --- whether dominance with some
coefficients decreasing  to zero is sufficient for prediction, we
have the following table of  questions and answers, where, in
fact, positive answers for $a_n$ are implied by positive answers
for $d_n$ and vice versa for the negative answers: \vskip 2mm
\begin{center}
\begin{tabular}{|c|c|c|c|c|c|}\hline
  $\E \bar d_n $ &  $\bar d_n \vphantom{\bar{\hat d}} $ &  $d_n$ &  $\E\bar a_n$& $\bar a_n$& $a_n$ \\\hline
  + & + &  $-$ & + & + & $-$\\\hline
\end{tabular}
\end{center}
However, if we take into account the conditions on the
coefficients,  we see some open problems left, and different
answers for $\bar d_n$ and $\bar a_n$ may be obtained. Following
is the table of conditions on dominance coefficients and answers
to the questions whether these conditions are sufficient for
prediction (coefficients bounded from below are included for the
sake of completeness).
\begin{center}
\begin{tabular}{|c|c|c|c|c|c|c|}\hline
                                                      & $\E \bar d_n $ &  $\bar d_n \vphantom{\bar{\hat d}} $ &  $d_n$ &  $\E\bar a_n$& $\bar a_n$& $a_n$\\\hline
 $\log c_n^{-1}=o(n)$                                 &  +             &  ?                                   &   $-$  &       +      &  ?         & $-$  \\\hline
 $\sum_{n=1}^\infty\frac{\log c_n^{-1}}{n^2}< \infty$ &  +             &  +                                   &   $-$ &         +     &  +         & $-$  \\\hline\hline
 $c_n\ge c>0$                                         &  +             &  +                                   &   $+$ & + & + &+ \\\hline
\end{tabular}
\end{center}
We know form Proposition~\ref{th:nodom} that the condition
$c_n\ge c>0$ for convergence in $d_n$ can not be improved; thus
the open problem left is to find whether  $\log c_n^{-1}=o(n)$ is
sufficient for prediction in $\bar d_n$ or at least in $\bar a_n$.

Another open problem is to find whether any conditions on
dominance coefficients are necessary for prediction; so far we
only have some sufficient conditions.
On the one hand, the
obtained results suggest that some form of dominance with decreasing
coefficients may be necessary for prediction, at least in the
sense of convergence of averages. On the other hand, the
condition~(\ref{eq:dom}) is uniform over all sequences which probably
is not necessary for prediction. As for prediction in the sense of
almost sure convergence, perhaps more subtle behavior of the
ratio $\frac{\mu(x_{1:n})}{\rho(x_{1:n})}$ should be analyzed,
since dominance with decreasing coefficients is not sufficient for
prediction in this sense.

\section{Preservation of the Predictive Ability under Summation with an Arbitrary Measure}\label{sec:sum}

Now we turn to the question whether, given a measure $\rho$ that
predicts a measure $\mu$ in some sense, the ``contaminated''
measure $(1-\eps)\rho+\eps\chi$ for some $0<\eps<1$ also predicts
$\mu$ in the same sense, where $\chi$ is an arbitrary measure.
Since most considerations are independent of the choice of $\eps$,
in particularly the results in this section, we set $\eps=\odt$
for simplicity. We define

\begin{definition}[Contamination]\label{def:cont}
By ``$\rho$ contaminated with $\chi$'' we mean
$\rho':=\odt(\rho+\chi)$.
\end{definition}

Positive results  can be obtained for  convergence in expected
average KL divergence. The statement of the next proposition in a
different form was mentioned in \cite{BRyabko:06,Hutter:06usp}.
Since the proof is simple we present it here for the sake of
completeness; it is based on the same ideas as the proof of
Theorem~\ref{th:dom}.

\begin{proposition}[\boldmath $\E\bar d\to 0$]\label{th:expaverklsum}
Let $\mu$ and $\rho$ be two measures on $\X^\infty$ and suppose
that $\rho$ predicts $\mu$ in expected average KL divergence. Then
so does the measure $\rho'=\odt(\rho +\chi)$ where $\chi$
is any other measure on $\X^\infty$.
\end{proposition}

\begin{proof}
\bqan
  0 \;\le\; \E \bar d_n(\mu,\rho')
  & = & \frac{1}{n}\E\sum_{t=1}^n \sum_{x_t\in\X} \mu(x_t|x_{<t})\log \frac {\mu(x_t|x_{<t})}{\rho'(x_t|x_{<t})}
  \;=\; \frac{1}{n}\E \log \frac {\mu(x_{1:n})}{\rho'(x_{1:n})}
\\
  & = & \frac{1}{n}\E \log \frac {\mu(x_{1:n})}{\rho(x_{1:n})}\frac{\rho(x_{1:n})}{\rho'(x_{1:n})}
  \;=\; \E\bar d_n(\mu,\rho)+\frac{1}{n}\E\log\frac{\rho(x_{1:n})}{\rho'(x_{1:n})},
\eqan
where the first term tends to 0 by assumption and the second term
is bounded from above by $\frac{1}{n}\log 2\rightarrow0$. Since
the sum is bounded from below by 0 we obtain the statement of the
proposition.\qedx
\end{proof}

Next we consider some  negative results. An example of measures
$\mu$, $\rho$ and $\chi$ such that $\rho$ predicts $\mu$ in
absolute distance (or KL divergence) but $\odt(\rho+\chi)$
does not, can be constructed similarly to the example from
\cite{Kalai:92} (of a measure $\rho$ which is a sum of
distributions arbitrarily close to $\mu$ yet does not predict it).
The idea is to take a measure $\chi$ that predicts $\mu$ much
better than $\rho$ on almost all steps, but on some steps  gives
grossly wrong probabilities.

\begin{proposition}[\boldmath $a\not\to 0$ and $d\not\to 0$]\label{th:nosumad}
There exist measures $\mu$, $\rho$ and $\chi$ such that $\rho$
predicts $\mu$ in absolute distance (KL divergence) but
$\odt(\rho+\chi)$ does not predict $\mu$ in absolute
distance (KL divergence).
\end{proposition}

\begin{proof}
Let $\mu$ be concentrated on the sequence $11111\dots$ (that is
$\mu(x_n=1)=1$ for any $n$), and let $\rho(x_n=1)={n\over n+1}$
with probabilities independent on different trials. Clearly,
$\rho$ predicts $\mu$ in both absolute distance and KL divergence.
Let $\chi(x_n=1)=1$ for all $n$ except on the sequence
$n=n_k=2^{2^k}=n_{k-1}^2$, $k\in\SetN$ on which
$\chi(x_{n_k}=1)=n_{k-1}/n_k=2^{-2^{k-1}}$. This implies that
$\chi(1_{1:n_k})=2/n_k$ and
$\chi(1_{1:n_k-1})=\chi(1_{1:n_{k-1}})=2/n_{k-1}=2/\sqrt{n_k}$. It
is now easy to see that $\odt(\rho+\chi)$ does not predict
$\mu$, neither in absolute distance nor in KL divergence. Indeed
for $n=n_k$ for some $k$ we have
\beqn
  \odt(\rho\!+\!\chi)(x_{n}=1|1_{<n})
  \;=\; \frac{\rho(1_{1:n})+\chi(1_{1:n})}{\rho(1_{<n})+\chi(1_{<n})}
  \;\leq\; \frac{1/(n\!+\!1) + 2/n} {1/n \;+\; 2/\sqrt{n}}
  \;\rightarrow\; 0.
\eeqn
\qedx
\end{proof}

For the (expected) average absolute distance the negative result also holds:

\begin{proposition}[\boldmath $\bar a\not\to 0$]\label{th:nosumavad}
There exist such measures $\mu$, $\rho$ and $\chi$ that $\rho$
predicts $\mu$ in  average absolute distance but
$\odt(\rho+\chi)$ does not predict $\mu$ in (expected)
average absolute distance.
\end{proposition}

\begin{proof}
Let $\mu$ be Bernoulli 1/2 distribution and let $\rho(x_n=1)=1/2$
for all $n$ (independently of each other) except for some sequence
$n_k$, $k\in\SetN$ on which $\rho(x_{n_k}=1)=0$. Choose $n_k$
sparse enough for $\rho$ to predict $\mu$ in the average absolute
distance. Let $\chi$ be Bernoulli  1/3. Observe that $\chi$
assigns non-zero probabilities to all finite sequences, whereas
$\mu$-a.s.\ from some $n$ on  $\rho(x_{1:n})=0$. Hence
$\odt(\rho+\chi)(x_{1:n})=\odt\chi(x_{1:n})$ and so
$\odt(\rho+\chi)$ does not predict $\mu$. \qedx
\end{proof}

Thus  for the question of whether predictive ability is preserved
when an arbitrary measure is added to the predictive measure, we
have the following table of answers.

\vskip 2mm
\begin{center}
\begin{tabular}{| c|c |c|c|c|c|c|}\hline
 $\E \bar d_n $ &  $\bar d_n \vphantom{\bar{\hat d}} $ &  $d_n$ &  $\E\bar a_n$& $\bar a_n$& $a_n$ \\\hline
 + & ? &  $-$ & $-$ & $-$ & $-$ \\\hline
\end{tabular}
\end{center}
\vskip 2mm

As it can be seen, there is one open question: whether this
property is preserved under almost sure convergence of the average
KL divergence.

It can be inferred from the example in
Proposition~\ref{th:nosumad} that contaminating   a predicting
measure $\rho$ with a measure $\chi$ spoils $\rho$ if $\chi$ is
better than $\rho$ on almost every step. It thus can be
conjectured that adding a measure can only spoil a predictor on
sparse steps, not affecting the average.

\begin{conjecture}[\boldmath $a\to 0$ implies $\bar a\to 0$]\label{con:sumad}
Suppose that a measure $\rho$ predicts a measure $\mu$ in absolute
distance. Then for any measure $\chi$ the measure
$\odt(\rho+\chi)$ predicts $\mu$ in average absolute
distance.
\end{conjecture}

As far as KL divergence is concerned we expect even a stronger
conjecture to be true, since limited KL divergence does not allow
a predicting measure to be (too close to) zero on any step.

\begin{conjecture}[\boldmath $\bar d\to 0$]\label{con:sumkl}
Suppose that a measure $\rho$ predicts a measure $\mu$ in average
KL divergence. Then for any measure $\chi$ the measure
$\odt(\rho+\chi)$ predicts $\mu$ in average KL divergence.
\end{conjecture}

\section{Miscellaneous}\label{sec:misc}

\paradot{Special cases of dominance with decreasing coefficients}
In Section~\ref{sec:dom} we have shown that Laplace's measure
$\rho_L$ for $\X=\{0,1\}$ dominates any Bernoulli i.i.d.\ measure
with linearly decreasing coefficients. It can also be shown that a
generalization of $\rho_L$ to a measure $\rho^k_L$ for predicting
any measure with memory $k$, for a given $k$,  dominates any such
measure with polynomially decreasing coefficients (namely,
$c_n^{-1}=\O(n^{|\X|^k}$). The measure $\rho_R$ from
\cite{BRyabko:88} for predicting any stationary measure was
constructed as a sum of $\rho^k_L$ with positive weights:
$\rho_R(x_{1..n})=\sum_{k=1}^{\infty}w_k\rho_L^k(x_{1..n})$. By
construction, $\rho_R$ dominates any finite memory measure with
polynomially decreasing coefficients. It is interesting to find
whether  $\rho_R$ (or any other measure which predicts all
stationary measures) dominates every stationary measure with some
subexponentially decreasing coefficients (or at least dominates
non-uniformly). Clearly, this is a special case of the general
open question --- whether some form of dominance with decreasing
coefficients is necessary for prediction.

\paradot{Special questions of summation of a predictor with arbitrary measures}
Although we know that adding a measure may spoil a predicting
measure, it may be that carefully selecting which groups of
measures to sum we can save all their predicting properties. One
of the interesting cases is Zvonkin-Levin \cite{Zvonkin:70}
universal semi-computable measure\footnote{In fact $\xi$ is not a
measure but only a semi-measure, but a semi-measure is sufficient
for making predictions and it will not affect our arguments
further.}
\beq\label{eq:xi}
\xi(x_{1:n})=\sum_{i\in\SetN} w_i\nu_i(x_{1:n})
\eeq
where ($\nu_i$), $i\in\SetN$ is the class of all lower
semi-computable semi-measures, and $w_i>0$. Since $\xi(A)\ge
w_i\nu_i(A)$ for any measurable set $A$ ($\xi$ dominates any
$\nu_i$ with a  constant $w_i$), it predicts all $\nu_i$ in the
sense of convergence in total variation,  in KL divergence and
absolute distance. The question is what else does it predict,
which other measures? Laplace's measure $\rho_L$ is computable,
and hence is present in the sum~(\ref{eq:xi}). We know that
$\rho_L$ predicts any Bernoulli i.i.d.\ measure, so we can ask
whether $\xi$, being a sum of $\rho_L$ and some other measures,
still predicts all Bernoulli measures. The predictor $\rho_R$ from
\cite{BRyabko:88} is computable and predicts all stationary
measures in average KL divergence (and average absolute distance).
Thus $\xi$ is a sum of $\rho_R$ and some other measure, and we can
ask whether it still predicts all stationary measures. (In
expected average KL divergence this follows from
Proposition~\ref{th:expaverklsum} as also pointed out in
\cite{Hutter:06usp}.)

\begin{conjecture}[\boldmath $a\to 0$ for i.i.d.\ and $\bar d\to 0$ for stationary]\label{conj:adb}
For every i.i.d.\ measure $\nu$, the measure $\xi$ as defined
in~(\ref{eq:xi}) predicts $\nu$  in absolute distance. For every
stationary measure $\nu$, the measure $\xi$ predicts $\nu$  in
average KL divergence.
\end{conjecture}

\paradot{Proof idea}
For the first question, consider any Bernoulli i.i.d.\ measure
$\nu$. From Conjecture~\ref{con:sumad} and from the fact that
$\rho_L$ predicts $\nu$ in absolute distance we conclude that
$\xi$ predicts $\nu$ in average absolute distance. Since $\nu$ is
Bernoulli i.i.d.\ measure, that is, probabilities on all steps are
equal and independent of each other, from any measure $\theta$
that predicts $\nu$ in average absolute distance we can make a
measure $\theta'$ which predicts $\nu$ in absolute distance as
follows
$\theta'(x_n=x|x_{<n})=\frac{2}{n}\sum_{t=n/2}^n\theta(x_t=x|x_{<n})$.
Moreover, the convergence speed of $\theta'$ to $\nu$ will be the
same as that of $\theta$ to $\nu$. But if $\xi$ is semi-computable
then so is $\xi'$, so that $\xi'$ is present in the sum
(\ref{eq:xi}). Since $\xi$ is not a better predictor than  $\xi'$,
adding $\xi$ to $\xi'$ can not spoil  the latter.

If $\nu$ is a stationary measure, then we known from
Proposition~\ref{th:expaverklsum} and that $\rho_R$ is computable
that $\xi$ predicts $\nu$ in expected average KL divergence
(absolute distance). Conjecture~\ref{con:sumkl} would also imply
that $\xi$ predicts $\nu$ in average KL divergence, and average
absolute distance.
\qed

\paradot{Other measures of divergence}
The last question we discuss is  criteria of prediction other than
introduced in Section~\ref{sec:def}. Apart form the measures of
divergence of probability measures that we considered we mention
also the following:
\begin{itemize}\itemindent=1ex
\item[($s$)] squared distance $s_n(\mu,\rho|x_{<n})=\sum_{x\in\X}(\mu(x_n=x|x_{<n})-\rho(x_n=x|x_{<n})^2$,
\item[($h$)]Hellinger distance $h_n(\mu,\rho|x_{<n})=\sum_{x\in\X}(\sqrt{\mu(x_n=x|x_{<n})}-\sqrt{\rho(x_n=x|x_{<n}\!}\;)^2$,
\end{itemize}
the average squared distance $\bar s_n$ and the average Hellinger
distance $\bar h_n$ are introduced analogously to $\bar a_n$ and
$\bar d_n$. It is easy to check that all negative results obtained
hold with respect to $s_n$ and $h_n$ as well. Positive results for
$s_n$ and $h_n$ follow from corresponding positive results for  KL
divergence $d_n$ and inequalities $s_n(\mu,\rho)\le d_n(\mu,\rho)$
and $h_n(\mu,\rho)\le d_n(\mu,\rho)$, see e.g.\
\cite[Lem.3.11]{Hutter:04uaibook}.
Expected absolute convergence $\E a_n\to 0$ (also called
convergence in the mean) and expected KL convergence $\E d_n\to 0$
may also be considered.

\section{Outlook and Conclusion}\label{sec:cocn}
In the present work we formulated and started to address the
question for which classes of measures sequence prediction is
possible. Towards this aim we defined the notion of dominance with
decreasing coefficients (a condition on local absolute continuity)
and found some forms of it which are sufficient for prediction. We
have also addressed the question which forms of predictive ability
are preserved under ``contamination'' of a predictor by an
arbitrary measure. Besides some more concrete open problems posed
in the corresponding sections, a  program for answering the
general questions formulated can be outlined as follows: We would
like to find some  conditions on dominance with decreasing
coefficients which are necessary and sufficient for prediction;
for those notions of prediction ability for which this is not
possible, more subtle behavior of the ratio
$\frac{\mu(x_{1:n})}{\rho(x_{1:n})}$ should be analyzed to
obtain conditions both necessary and sufficient for prediction.
This should give rise to an abstract characterization of classes
of measures for which a measure satisfying such conditions  for
all measures in the class exists; that is, to a description of
classes of measures for which prediction is possible. It is
expected that such characterization will naturally lead to a
construction of a predictor as well~--- perhaps in form of a
Bayesian integral. The latter conjecture also encourages studying
the question of ``contamination'' of a predictor with arbitrary
measures. The next step will be to extend this approach to the
task of active learning \cite{Hutter:04uaibook,Ryabko:06actopt}.


\begin{small}

\end{small}

\end{document}